# Joint-YODNet: A Light-weight Object Detector for UAVs to Achieve Above 100fps

Vipin Gautam*, Shitala Prasad*, and Sharad Sinha

School of Mathematics and Computer Science
Indian Institute of Technology Goa
{shitala, vipin2113106, sharad}@iitgoa.ac.in

**Abstract.** Small object detection via UAV (Unmanned Aerial Vehicle) images captured from drones and radar is a complex task with several formidable challenges. This domain encompasses numerous complexities that impede the accurate detection and localization of small objects. To address these challenges, we propose a novel method called Joint-YODNet for UAVs to detect small objects, leveraging a joint loss function specifically designed for this task. Our method revolves around the development of a joint loss function tailored to enhance the detection performance of small objects. Through extensive experimentation on a diverse dataset of UAV images captured under varying environmental conditions, we evaluated different variations of the loss function and determined the most effective formulation. The results demonstrate that our proposed joint loss function outperforms existing methods in accurately localizing small objects. Specifically, our method achieves a recall of 0.971, and a F1Score of 0.975, surpassing state-of-the-art techniques. Additionally, our method achieves a mAP@.5(%) of 98.6, indicating its robustness in detecting small objects across varying scales.

## 1 Introduction

Unmanned aerial vehicles (UAVs) have garnered significant popularity across various domains, encompassing surveillance, search and rescue operations, and environmental monitoring [21,27,1]. The detection of small objects, such as vehicles or pedestrians, within UAV images and radar scans holds the utmost importance for numerous applications. However, accomplishing this task is inherently challenging due to a multitude of factors including limitations in resolution, object occlusion, background clutter, low contrast, motion blur, complexities in data annotation, and real-time processing requirements.

Despite these challenges, there has been significant progress in recent years in the development of small object detection algorithms for UAV/aerial imaging [23,12]. These algorithms typically use a combination of techniques, such as image segmentation, feature extraction, and machine learning, to identify and classify small objects. Some of the most promising approaches include deep learning methods, which have been shown to be very effective at extracting features from aerial images.

---

* These authors contributed equally to this work.

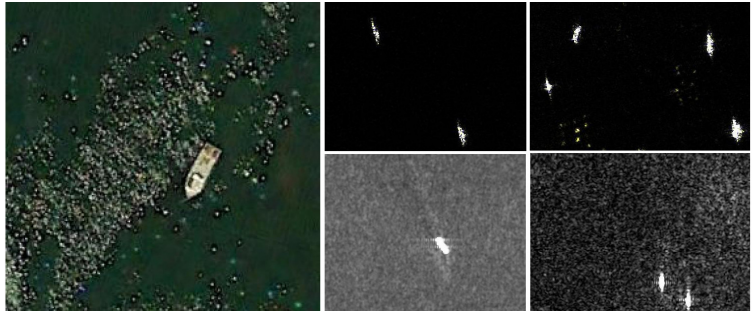

Fig. 1: Demonstration of Detection challenges in images obtained through UAV.

Advancements in deep learning have improved object detection methods used in various applications, including aerial imaging [7,9]. However, these methods face challenges in detecting small objects in high-resolution aerial images. Detecting objects in aerial images presents several challenges due to the high resolution and presence of tiny objects. Challenges include information loss from rescaling, low tolerance to bounding box shifts, and noisy feature representations [3]. A slight shift in the bounding box can lead to false positives and a decrease in Intersection over Union (IoU) [3]. In the existing literature, various object detection methods have been proposed with different strategies to improve system performance. These strategies include enhancing deep network architectures [20], introducing new loss functions [16], and proposing innovative learning approaches [17]. Among these, the use of dedicated loss functions has shown significant improvements in overall performance, motivating the focus of this paper on introducing a new loss function.

Specifically, we propose a joint loss function tailored for small object detection in aerial views. This joint loss is integrated into a deep model, enhancing the learning capability of the network compared to conventional training methods. The optimization of gradients and convergence leads to higher detection accuracy. To further improve feature representation, we incorporate the Omni-dimensional dynamic convolution (ODConv) [10], which enhances overall performance. Furthermore, we explore the training process using a smaller dataset, making our system more practical and easier to deploy in real-world scenarios.

In this study, we focus on detecting ships and vessels using drone imagery, see Fig. 1. These objects play crucial roles in maritime surveillance, security, and navigation applications. However, detecting these relatively large objects in UAV images and radar scans can be challenging due to varying scales, occlusion, and cluttered backgrounds [24,26]. Small ships, in particular, can be easily confused with ocean streams due to ship motion. Additionally, occlusion from other ships, structures, or environmental factors, along with background clutter such as waves or coastal features, further complicates the detection process.

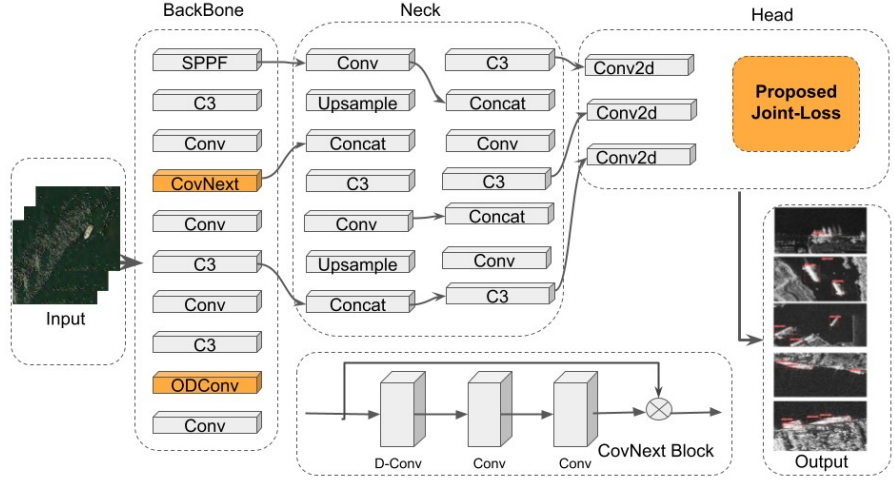

Fig. 2: Joint-YODNet Structure

## 2    Methodology

This section presents a comprehensive and advanced overview of our proposed light-weight small object detector called Joint-YODNet (Joint-loss based YOLO Omni-dimensional Dynamic Convolution Network) and discusses its various modules in detail. Our innovative object detector enables us to efficiently extract and accurately classify multiple small objects.

### 2.1    You Only Look Once (YOLO)

YOLO version 5 (YOLOv5), a lightweight CNN-based object detection model, is widely recognized for its efficiency [8]. The model comprises three main components: the Backbone, the Neck, and the Head. The Backbone utilizes a convolutional neural network (CNN) to extract high-quality features from the input image. Notably, it incorporates the CSP module (C3), inspired by the CSPNet design [22]. The Neck consists of additional convolutional layers that capture intricate details and spatial information in the feature maps. The Detection Head utilizes the processed feature maps to perform the final steps of object detection, including bounding box prediction and class probability estimation.

Loss function consists of three components: objectness loss, localization loss, and classification loss. These components are combined using weights to optimize the model for accurate object detection, precise bounding box prediction, and correct classification during training. Overall loss function can be expressed as:

$$L_{loss} = \lambda_1 L_{cls} + \lambda_2 L_{obj} + \lambda_3 L_{loc} \qquad (1)$$

Binary cross entropy (BCE) loss is used for classification loss $L_{cls}$ and object loss $L_{obj}$ while for localization loss $L_{loc}$ CIoU loss is used. Detailed discussion is done in section 2.3 which demonstrates the different types of localization loss functions used for bounding box regression.

## 2.2   ODConvNeXt

We present YOLOv5-ODConvNeXt, a variant of YOLOv5 that incorporates the CovNext [15] and ODConv [10] modules within the backbone network, as shown in Fig. 2. SAR image analysis poses several challenges such as speckle noise, complex texture, limited training data, scale/resolution variations, shadow/layover effects, and data interpretation. To address these challenges, we leverage the Cov-Next and ODConv modules to enhance the feature maps obtained by YOLOv5-ODConvNeXt [2].

Additionally, we introduce a "Joint Loss" function specifically designed to improve the localization accuracy of bounding boxes and accelerate convergence. The effectiveness of the joint loss function in enhancing localization ability is evaluated through a comprehensive set of experiments, both quantitative and qualitative, as discussed in Section 4.

## 2.3   Loss Functions for BBox

Bounding box (BBox) regression is a key technique in object detection that predicts the location of target objects using rectangular BBoxes. It aims to improve the accuracy of the predicted BBox.

To achieve this, the regression process uses loss functions based on the Intersection over Union (IoU), which measures the overlap between the predicted BBox and the ground truth BBox. The IoU is calculated as the ratio of the intersection area between the ground truth and predicted BBoxes to their union:

$$IoU = \frac{|GT \cap PD|}{|GT \cup PD|} \qquad (2)$$

The IoU loss function is effective when there is an overlap between the predicted ($PD$) and ground truth ($GT$) BBoxes. However, it struggles to produce meaningful gradients and slow convergence when there is no overlap between the BBoxes.

**Generalized Intersection over Union** (GIoU) [19] loss maximizes the overlap between the predicted and ground truth BBoxes by gradually adjusting the size of the predicted box. It is particularly effective when the boxes initially do not overlap. The GIoU formula is defined as follows:

$$GIoU = IoU - \frac{|C - (GT \cup PD)|}{|C|} \qquad (3)$$

Where C represents the minimum bounding box that encompasses both the predicted (PD) and ground truth (GT) BBoxes. It acts as a penalty term, guiding the predicted BBox towards the target ground truth BBox. The GIoU loss outperforms Mean Squared Error (MSE) loss and IoU loss in terms of precision. While it addresses the issue of vanishing gradients in non-overlapping scenarios, it may have slower convergence and less accurate regression for boxes with extreme aspect ratios.

**Distance IoU** (DIoU) [31] is a measure of the normalized distance between the center points of the predicted and ground truth BBoxes. By incorporating distance information, it enables faster convergence and more precise regression.

$$DIoU = IoU - \frac{d^2}{c^2} \qquad (4)$$

In this formula, '$d$' represents the Euclidean distance between the center points of the predicted and ground truth BBoxes, while '$c$' denotes the diagonal length of the smallest enclosing box that covers both BBoxes. The inclusion of distance information in the loss function enhances optimization by enabling faster convergence. It also improves the accuracy of regression, resulting in better localization of objects in object detection tasks.

**Complete Intersection over Union** (CIoU) [32] loss incorporates three essential geometric factors: overlap area, distance, and aspect ratio. CIoU loss is a versatile approach to BBox regression, surpassing both GIoU and DIoU. However, when the aspect ratio of the ground truth BBox matches that of the predicted BBox, CIoU degenerates to DIoU.

**Efficient IOU** [29] addresses limitations of traditional IOU loss by incorporating additional components to reflect the closeness between BBoxes and improve convergence. The Efficient IOU loss consists of three terms: IOU loss ($L_{iou}$), distance loss ($L_{dis}$), and aspect ratio loss ($L_{asp}$). By combining these terms, the Efficient IOU loss enhances the training efficiency and achieves improved performance. Retaining the positive effects of CIoU loss, Efficient IOU demonstrates the potential for further improvement in neural network training. Thus, it is defined as:

$$L_{eiou} = 1 - IOU + \frac{\rho^2(b, b^{GT})}{(w^c)^2 + (h^c)^2} + \frac{\rho^2(w, w^{GT})}{(w^c)^2} + \frac{\rho^2(h, h^{GT})}{(h^c)^2} \qquad (5)$$

where $h^w$ and $h^c$ are the width and height of the smallest enclosing box covering the two boxes. Variables $b$ and $b^{GT}$ are the center of the predicted and ground truth BBox.

**Proposed Joint Loss** We propose the "**Joint Loss**", an enhanced approach that combines multiple loss components to address existing limitations. The Joint Loss is computed using a formula involving coefficients $\alpha, \beta, \gamma$, and $\eta$, determined through empirical tests: $\alpha = 0.1$, $\beta = 0.1$, $\gamma = 0.1$, $\eta = 0.7$.

$$L_{joint} = \alpha L_{ciou} + \beta L_{diou} + \gamma L_{giou} + \eta L_{eiou} \qquad (6)$$

Compared to a single loss, a joint loss function combines multiple components, enabling the model to optimize for multiple objectives simultaneously. It improves overall performance, captures different aspects of the problem, and addresses challenges through specific loss components. The joint loss helps overcome the gradient vanishing problem and facilitates generalization to unseen data. It allows customization and fine-tuning, incorporating domain-specific knowledge for better results. Mathematically, the joint loss is denoted as $L_{joint}$ and consists of individual loss components $L_{ciou}$, $L_{diou}$, $L_{giou}$, $L_{eiou}$. The gradients of the joint loss are computed by summing the gradients of the individual components, as shown in Eq. 7. Binary cross entropy (BCE) is used for classification. In the following section, we present the experimental setup and evaluation of the proposed loss function.

$$\Delta L_{joint} = \alpha \Delta L_{ciou} + \beta \Delta L_{diou} + \gamma \Delta L_{giou} + \eta \Delta L_{eiou} \qquad (7)$$

## 3   Implementation and Datasets

The hardware setup consists of 2 x Intel Xeon Gold 6248 processors, each with 20 cores running at a clock speed of 2.5GHz, and an NVIDIA DGX A100 GPU. All experiments are conducted using the Python language and the PyTorch framework. The proposed model is optimized using Stochastic Gradient Descent (SGD) with an initial learning rate of 0.01, a momentum factor of 0.937, and weight decay set to 0.0005. During training, a batch size of 32 is used, and the training process is carried out for a total of 500 epochs. The network input size used in this work is 640x640. Additionally, the Intersection over Union (IoU) threshold is set to 0.5 for the experiments.

### 3.1   Evaluation Criterion

To evaluate the proposed method, we used standard performance metrics, including precision (P), recall (R), F1 measure (F1), and mean average precision (mAP). The mAP is calculated as the average of the average precision (AP) values for each category, which are obtained from the Precision-Recall (PR) curve. Precision and recall are computed using the formulas:

$$P = \frac{TP}{TP+FP}, \ R = \frac{TP}{TP+FN}, \ F1 = 2 \cdot \frac{P \cdot R}{P+R} \qquad (8)$$

where TP represents true positives, FP represents false positives, and FN represents false negatives. Frames per second (FPS) has been used to evaluate the real-time application of the model. We base FPS calculation on the formula provided by the official repository created by Ultralytics [8]. The formula used for FPS calculation can be defined as.

$$FPS = \frac{1000}{P+I+NMS} \qquad (9)$$

where $P$, $I$, and $NMS$ are preprocessing, inference, and Non Max suppression time, respectively.

### 3.2   Datasets Description

For evaluating the proposed methods, we utilized two well-known benchmark datasets for Aerial Image Detection: SAR Ship Detection and Aerial Ship Detection datasets. These datasets are widely recognized and present significant challenges in the field of aerial image for small object detection.

**SAR Ship Detection Dataset**   For the SAR Ship Detection (SAR-SD) dataset, we utilized the publicly available SSDD dataset [11]. This dataset contains 1160 SAR images with 2540 ship instances of varying resolutions (1 to 15 meters). The ships in this dataset are small, with some being only tens of pixels in size. We divided the SAR-SD dataset into a training set of 641 images, 271 for the validation set, and a separate test set of 232 images. Our focus was to assess the generalization capability of the proposed method using this dataset.

**Aerial Ship Detection Dataset**   The Aerial Ship Detection (ASD) dataset, sourced from Roboflow [5], comprises 1395 JPEG aerial images. The images have a resolution of 600x400 pixels. We used an official split of a training set of 1224, a validation set of 113, and a test set of size 58 images. To ensure uniformity, the images underwent preprocessing steps including auto-orientation and resizing to 640x640 pixels. Data augmentation techniques were applied, including a noise augmentation called salt and pepper, which introduced up to 5% noise through pixel manipulation within the bounding boxes. These augmentations were implemented to enhance the robustness and generalizability of the ship detection model trained on this dataset.

## 4   Experimental Results

This section presents the experimental analysis conducted on the aforementioned public datasets. The results are comprehensively examined, encompassing both qualitative and quantitative aspects. We compare our method with existing state-of-the-art (SOTA) approaches, followed by conducting ablation studies to evaluate the impact of different components in our proposed method. Additionally, we assess the computational speed of our method to gauge its efficiency.

### 4.1   Comparison with SOTA Methods:

Our proposed method is evaluated and compared to several SOTA networks for SAR-SD dataset, as summarized in Table 1. The results demonstrate that our method achieves competitive performance compared to existing networks. YOLOv5-ODConvNeXt [2] achieves remarkable results, with a precision of 0.971, recall of 0.96, and F1Score of 0.965. It also achieves an mAP@.5(%) of 98.10 and an mAP@:.95(%) of 72.70. However, FIERNet [24] and CR2A-Net [26] achieve the highest precision scores of 0.98. Our proposed Joint-YODNet achieves a precision of 0.979, demonstrating excellent performance comparable to FIERNet and CR2A-Net. In terms of recall, F1Score, and mAP@.5(%), Joint-YODNet achieves a recall of 0.971, an F1Score of 0.975, and a mAP@.5(%) of 98.6, outperforming YOLOv5-ODConvNeXt and other compared methods. SOTA methods, including SSD [14], Faster R-CNN [18], YOLOv5 [8], RetinaNet [13], DDNet [30], Quad-FPN [28], SAR-ShipNet [4], and HA-SARSD [25], exhibit varying levels of performance in terms of precision, recall, F1Score, mAP@.5(%), and mAP@.5:.95(%).

In summary, our proposed method shows competitive performance compared to existing state-of-the-art networks. It achieves high precision comparable to FIERNet and CR2A-Net, while surpassing YOLOv5-ODConvNeXt and other methods in terms of recall and F1Score. Furthermore, our method demonstrates strong performance in terms of mAP@.5(%) and mAP@.5:.95(%).

Table 1: State-of-the-art comparison with existing networks on SAR-SD.

| Networks | P | R | F1 | mAP@.5(%) | mAP@.5:.95(%) |
|---|---|---|---|---|---|
| SSD [14] | 0.846 | 0.811 | 0.828 | – | – |
| Faster R-CNN [18] | 0.871 | 0.856 | 0.863 | – | – |
| YOLOv5 [8] | 0.964 | 0.897 | 0.929 | 95.2 | – |
| RetinaNet [13] | 0.901 | 0.891 | 0.896 | – | – |
| DDNet [30] | 0.931 | 0.912 | 0.921 | – | – |
| Quad-FPN [28] | 0.90 | 0.957 | 0.925 | 95.20 | – |
| SAR-ShipNet [4] | 0.95 | 0.763 | 0.847 | 89.08 | – |
| FIERNet [24] | **0.98** | 0.879 | 0.927 | 94.14 | – |
| CR2A-Net [26] | **0.98** | 0.878 | 0.927 | 89.8 | – |
| HA-SARSD [25] | 0.97 | 0.920 | 0.944 | 97.0 | – |
| YOLOv5-ODConvNeXt [2] | 0.971 | <u>0.96</u> | <u>0.965</u> | <u>98.10</u> | **72.70** |
| *Joint-YODNet* | <u>0.979</u> | **0.971** | **0.975** | **98.6** | <u>72.6</u> |

**Bold**: best; <u>Underlined</u>: second-best results. The dash represents results not found.

### 4.2   Ablation study

**Evaluation on SAR-SD**  The Table 2 compares different loss variations for small object detection on the SAR-SD dataset. The baseline method, YOLOv5-ODConvNeXt, achieved a precision of 0.971, recall of 0.96, F1Score of 0.965, mAP@.5(%) of 98.10, and mAP@.5:.95(%) of 72.70. The EIoU method improved precision (0.984) and F1Score (0.976), achieving the highest values in these metrics. The SIoU method also performed well, achieving a precision of 0.979 and an F1-score of 0.974.

The combination of EIoU and SIoU, denoted as EIoU + SIoU, resulted in a precision of 0.975, recall of 0.969, F1Score of 0.972, mAP@.5(%) of 98.50, and mAP@.5:.95(%) of 72.70.

Our Joint-YODNet achieved a precision of 0.979, recall of 0.974, F1Score of 0.976, mAP@.5(%) of 98.60, and mAP@.5:.95(%) of 72.60. These results show that our method outperforms the baseline YOLOv5-ODConvNeXt method and performs at par with the EIoU and SIoU methods.

Overall, the results in Table 2 show that our proposed method is a promising approach for small object detection in SAR images. It achieves superior performance compared to the baseline method and performs at par with the SOTA methods. Additionally, our method is computationally efficient and has a lightweight architecture, making it well-suited for real-time applications and deployment on resource-constrained edge devices (discussed in 4.4).

Table 2: Comparison of various bounding box regression losses on SAR-SD.

| Loss Variations | P | R | F1 | mAP@.5(%) | mAP@.5:.95(%) |
|---|---|---|---|---|---|
| YOLOv5-ODConvNeXt [2] | 0.971 | 0.96 | 0.965 | 98.10 | <u>72.70</u> |
| EIoU [29] | **0.984** | 0.968 | **0.976** | **98.60** | **72.80** |
| SIoU [6] | 0.979 | 0.969 | <u>0.974</u> | 98.40 | 72.10 |
| EIoU + SIoU | 0.975 | <u>0.969</u> | 0.972 | <u>98.50</u> | 72.70 |
| *Joint-YODNet* | <u>0.979</u> | **0.974** | **0.976** | **98.60** | 72.60 |

**Bold**: best; <u>Underlined</u>: second-best results.

**Evaluation on ASD Dataset** The experimental results presented in Table 3 illustrate the performance comparison of different loss variations on the ASD dataset. Notably, our proposed method exhibits promising results, showcasing competitive performance across various metrics. Our method achieves a precision of 0.84, outperforming other loss variations, including YOLOv5-ODConvNeXt (0.61), EIoU (0.62), and SIoU (0.62). Furthermore, our approach achieves a recall of 0.63, surpassing all other variations.

In terms of F1-score, our method achieves a value of 0.720, indicating a balanced trade-off between precision and recall. This outperforms other loss variations, including YOLOv5-ODConvNeXt (0.705), EIoU (0.699) and SIoU (0.707). Moreover, our method demonstrates superior performance in terms of mAP-@.5(%), and mAP@.5:.95(%) metrics. Our method achieves a mAP@.5(%) of 70.30, surpassing other variations, including YOLOv5-ODConvNeXt (68.90), EIoU (66.60), and SIoU (68.10). Similarly, our method achieves a mAP@.5:.95(%) of 31.20, showcasing competitive performance compared to other loss variations, such as YOLOv5-ODConvNeXt (31.40), EIoU (31.10), and SIoU (29.10).

Table 3: Comparison of various bounding box regression losses on ASD dataset.

| Loss Variations | P | R | F1 | mAP@.5(%) | mAP@.5:.95(%) |
|---|---|---|---|---|---|
| YOLOv5-ODConvNeXt [2] | <u>0.85</u> | 0.61 | 0.705 | <u>68.90</u> | **31.40** |
| EIoU [29] | 0.81 | <u>0.62</u> | 0.699 | 66.60 | 31.10 |
| SIoU [6] | 0.83 | 0.62 | 0.707 | 68.10 | 29.10 |
| EIoU + SIoU- Combined | **0.88** | 0.60 | <u>0.716</u> | 68.20 | 29.90 |
| *Joint-YODNet* | 0.84 | **0.63** | **0.720** | **70.30** | <u>31.20</u> |

**Bold**: best; <u>Underlined</u>: second-best results.

### 4.3   Qualitative Study

A qualitative analysis was undertaken to compare the detection performance of the proposed Joint-YODNet with that of YOLOv5-ODConvNeXt, as depicted in Fig. 3. The analysis involved examining visual results derived from images showcasing both the ground truth ship instances and the detections generated by the two methods. Specifically, (a) denotes the ground truth bounding boxes, (b) represents the detection results obtained from YOLOv5-ODConvNeXt, and (c) corresponds to the bounding box detections obtained through the proposed method. The visual comparison vividly demonstrates the superior performance of our proposed method in accurately localizing and recognizing ship instances despite their varying small sizes.

Conversely, the detections generated by YOLOv5-ODConvNeXt (a) exhibit certain limitations. Some instances are either missed or incorrectly identified, resulting in lower precision and recall, as evidenced in Fig. 3. These disparities in detection quality further emphasize the advantages of our proposed method in real-life scenarios.

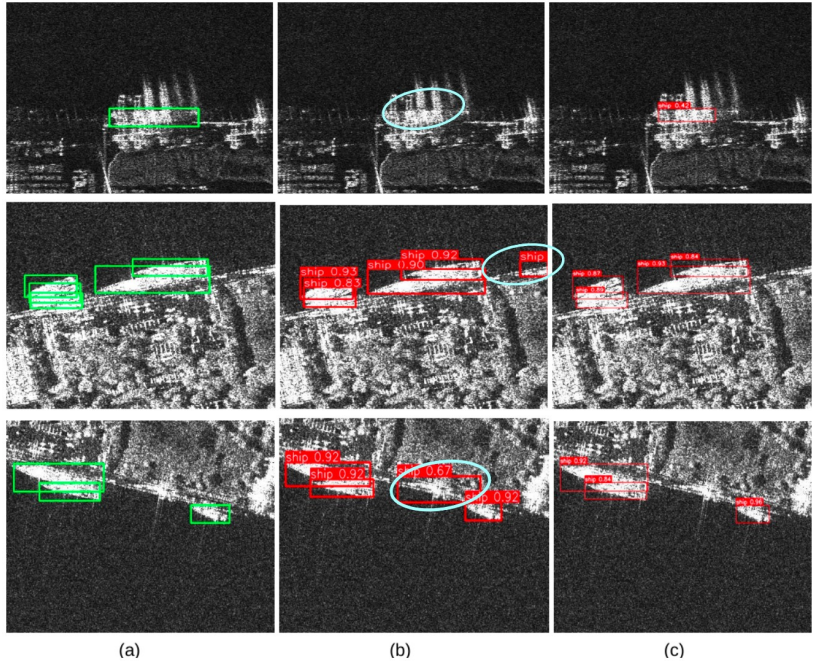

Fig. 3: Comparing Object Detection Performance on the SAR-SD Dataset: (a) Ground Truth Labels, (b) Detections by YOLOv5-ODConvNeXt, and (c) Detections achieved through the proposed method.

### 4.4 Computational Analysis

Joint-YODNet not only achieves superior performance but also demonstrates remarkable computational efficiency, with a high frame rate (FPS) of 136 (used Eq. 9). This makes it ideal for real-time applications that require timely and accurate ship detection. Additionally, our method features a lightweight architecture, making it compatible with resource-constrained edge devices. This enhances its versatility for deployment in various practical scenarios. Joint-YODNet has a parameter size of 6.99M and a memory size of 14.4MB.

## 5   Conclusion

In this study, we proposed a novel method for small object detection in UAV imagery, utilizing a tailored joint loss function. Our approach, Joint-YODNet outperforms existing methods, achieving superior precision, recall, F1Score, and mAP@.5 scores, demonstrating its robustness across object scales. The qualitative analysis further validates the effectiveness of our method in real-life scenarios. Our research contributes to the field by addressing the challenges of small object detection in UAV imagery and enabling accurate localization and recognition. The results highlight its potential in surveillance, object tracking, and environmental monitoring in UAV-based systems. Future research can focus on refining the joint loss function and integrating additional techniques for further enhancement.